\documentclass[letterpaper, 10 pt, conference]{ieeeconf}  
\usepackage{graphicx} 
\usepackage{xcolor}
\usepackage{amsmath}
\usepackage{amsfonts}
\usepackage{algorithm}
\usepackage{algorithmic}

\IEEEoverridecommandlockouts                              

\title{\LARGE \bf
Distributed Surface Inspection via Operational Modal Analysis by a Swarm of Miniaturized Vibration-Sensing Robots
}

\author{Thiemen Siemensma$^{1}$, Niels de Boer$^{1}$, and Bahar Haghighat$^{1}$
\thanks{$^{1}$All authors are with the Faculty of Science and Engineering,
        University of Groningen, 9747 AG Groningen, The Netherlands
        }
}

\begin{document}

\maketitle
\thispagestyle{empty}
\pagestyle{empty}

\begin{abstract}
Robot swarms offer the potential to serve a variety of distributed sensing applications. An interesting real-world application that stands to benefit significantly from deployment of swarms is structural monitoring, where traditional sensor networks face challenges in structural coverage due to their static nature. This paper investigates the deployment of a swarm of miniaturized vibration sensing robots to inspect and localize structural damages on a surface section within a high-fidelity simulation environment. In particular, we consider a 1 m × 1 m × 3 mm steel surface section and utilize finite element analysis using Abaqus to obtain realistic structural vibration data. The resulting vibration data is imported into the physics-based robotic simulator Webots, where we simulate the dynamics of our surface inspecting robot swarm. We employ (i) Gaussian process estimators to guide the robots' exploration as they collect vibration samples across the surface and (ii) operational modal analysis to detect structural damages by estimating and comparing existing and intact structural vibration patterns. We analyze the influence of exploration radii on estimation uncertainty and assess the effectiveness of our method across 10 randomized scenarios, where the number, locations, surface area, and depth of structural damages vary. Our simulation studies validate the efficacy of our miniaturized robot swarm for vibration-based structural inspection.
\end{abstract}

\section{Introduction}
\label{sec:introduction}
Swarms of miniaturized robots hold great potential for distributed sensing and monitoring tasks \cite{Dorigo2021SwarmFuture}. Applications arise in agriculture, defense, space, and inspection of machinery \cite{Albani2017FieldSwarms, Dawkins2018DeploymentChallenge,Nguyen2019Swarmathon:Exploration, Correll2009FromMachinery}. A particularly interesting application area is Structural Health Monitoring (SHM), where high labor costs and limited human reliability have driven the development of automated damage detection methods \cite{Farrar2007AnMonitoring, Gharehbaghi2022APerspectives, Phares2004RoutineReliability}.
Traditionally, these methods rely on static sensor networks, whose effectiveness depends heavily on sensor density and placement \cite{Mustapha2021SensorReview}. Despite sensor placement optimization efforts, static sensors remain limited in their ability to detect localized damages when structural coverage is sparse \cite{Bigoni2020SystematicStates, Ostachowicz2019OptimizationReview}. More recently, robotic solutions have emerged to address these limitations \cite{Lattanzi2017ReviewSystems}. In particular, robotic systems have been successfully deployed for visual assessment of infrastructure such as wind turbines, bridges, storage tanks, and ship hulls \cite{Liu2022ReviewTurbines, Oh2009BridgeVision, Acar2019AutonomousInspection, Ahmed2015DesignInspection}.
While effective for detecting surface-level defects, visual methods are inherently limited in capturing subsurface or internal damage. 
Within SHM, a more comprehensive structural assessment is achieved through vibration sensing, which can reveal damages prior to visible deterioration through detecting changes in vibration response \cite{Avci2021AApplications,Almeida2014AnLocalization, Pandey1991DamageShapes, Dawari2013StructuralDifferences}.
\begin{figure}[t!]
    \centering
    \includegraphics[width=0.94\columnwidth]{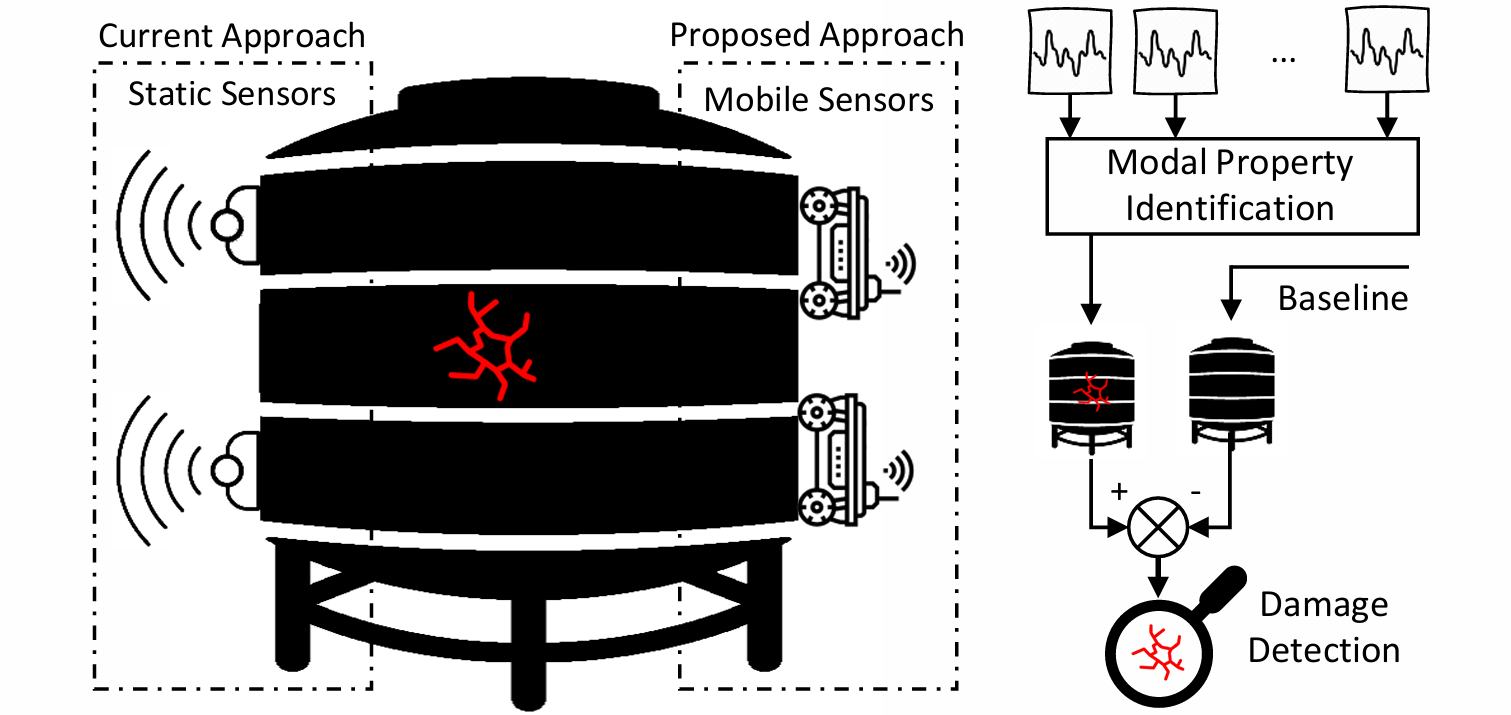}
    \caption{Vibration-based structural inspection approaches illustrated using a water tank sketch as an example. Vibration signals are recorded and used to extract modal properties via operational modal analysis \cite{Mostafaei2025StatePerspectives}. These properties are compared against a baseline undamaged state to detect and localize structural damages. While conventional approaches rely on static sensor networks, we propose the use of a miniaturized robot swarm realizing mobile sensors to enhance the inspection's spatial coverage.}
    \label{fig:shm_example}
\end{figure}

Studies that recognize the potential of integrating vibration sensors with mobility include \cite{Zhu2009AMonitoring,Guo2009AMonitoring}, where wheeled robots are used to identify structural properties. However, their reliance on single-robot systems with relatively large hardware can limit coverage effectiveness and decrease damage detection accuracy due to added robot weight.
This highlights the potential of miniaturized robot swarms in SHM, where miniaturized light-weight hardware and efficient coverage strategies are enablers for non-intrusive and efficient structural assessment \cite{Brambilla2013SwarmPerspective}.
Several studies have explored the use of vibration sensing within swarms for tasks such as binary indication, source localization, and robot aggregation \cite{Siemensma2024CollectiveInspection, Haghighat2022AnRobots, Hao2023ControllingRobots}. 
However, these methods do not incorporate any realistic assessment of structural properties, limiting their applicability to real world structural inspection tasks. To advance deployment of swarm robotics toward these real world applications, integration with established SHM methodologies for inspection is necessary ( Figure~\ref{fig:shm_example}).

A common established SHM method is Operational Modal Analysis (OMA), which identifies modal parameters that, when compared to baseline (undamaged) data, enable to infer the presence, location, and severity of structural damage \cite{Almeida2014AnLocalization, Mostafaei2025StatePerspectives, Dawari2013StructuralDifferences,Rytter1993VibrationalStructures}.
Modal parameters include natural frequencies, damping ratios, and mode shapes, the latter representing the spatial distribution of vibration amplitudes.
Modal identification methods include the Peak Picking (PP) method, which identifies peaks in the frequency response at sensor locations \cite{Zimmerman2008AutomatedSystems}. Employing time domain analysis, the Stochastic Subspace Identification  (SSI) algorithm estimates system dynamics and noise directly from time-series data \cite{Peeters1999Reference-BasedAnalysis}. Frequency domain analysis can provide more robustness to noise; the Frequency Domain Decomposition (FDD) method applies Singular Value Decomposition (SVD) to structural frequency response \cite{Brincker2001ModalDecomposition}. By focusing on dominant eigenvalues and eigenvectors, information unrelated to modal parameters is filtered out. 
While these OMA methods have been successfully applied to static sensor networks \cite{Zimmerman2008AutomatedSystems}, they do not inherently provide global and continuous estimates of structural properties. Achieving such estimates requires integration with mobile sensors that employ effective navigation and spatial field estimation strategies.
Within multi-robot systems, Bayesian methods, particularly Gaussian Processes (GP), have proven to be robust and well-established tools for 
exploration and scalar field estimation tasks \cite{Benevento2020Multi-RobotFields, Luo2018AdaptiveProcesses, Jang2020Multi-RobotProcess, Carron2015Multi-agentsRegression}. Robots use GP models to identify and sample uncertain areas, thereby improving the overall estimation of the underlying field. 
Extensions to the GP framework have been proposed to support online learning and planning \cite{Ma2017InformativeProcesses,Mishra2018OnlineProcesses}, and coordination among multiple robots \cite{Jang2020Multi-RobotProcess,Pillonetto2017DistributedApproximations}. 


In this work, we integrate established OMA and field estimation techniques. In particular, we propose and validate a method for detection and localization of damages using a swarm of vibration-sensing robots. 
The primary contribution of our work is an implementation of the FDD analysis combined with GP exploration and field estimation, enabling a mobile OMA scheme. Our implementation is tailored to minimize communication, computation, and data storage, with these reductions being tunable to match the capabilities of resource-constrained robots.
A secondary contribution of our work is a high-fidelity simulation framework that combines realistic vibration data from Abaqus \cite{DassaultSystemesWww.3ds.com/products/simulia/abaqus} with physics-based robot dynamics simulations in Webots \cite{WebotsWww.cyberbotics.com} built around a calibrated model of an existing miniaturized vibration-sensing robot (see Figure \ref{fig:three-in-one-column_robot}). This simulation environment enables future studies of more complex structures and damage scenarios, supporting future development of inspection methods and their transferability to real hardware experiments. To the best of our knowledge, this work is the first to investigate and demonstrated applicability of robot swarms to realistic vibration-based structural inspection.
\section{Problem Definition}
\label{sec:problem-definition}
We consider a swarm of miniaturized wheeled robots operating on a metallic structure, tasked with detecting and localizing an unknown number of structural damages using vibration-based sensing. These vibrations arise from ambient excitations such as wind or traffic loads \cite{Zimmerman2008AutomatedSystems,Deraemaeker2008Vibration-basedEnvironment}.
The robots follow a two-phase strategy. In the \textit{exploration} phase, they collect vibration data across the surface in an informative manner. In the \textit{estimation} phase, the acquired data is processed in a decentralized fashion to produce a continuous and spatially distributed damage estimate.

In our study, structural damage is modeled as a localized reduction in stiffness, which alters the dynamic response of the structure and manifests changes in mode shape curvature \cite{Almeida2014AnLocalization,Pandey1991DamageShapes,Dawari2013StructuralDifferences}. The robots compare the curvature of the current vibration modes to a known baseline undamaged state and aggregate these deviations to infer damage. 

\section{Simulation Framework}
\label{sec:methods}
\begin{figure}[t!]
  \centering
  \begin{minipage}[b]{0.32\columnwidth} 
    \centering
    \includegraphics[width=\linewidth]{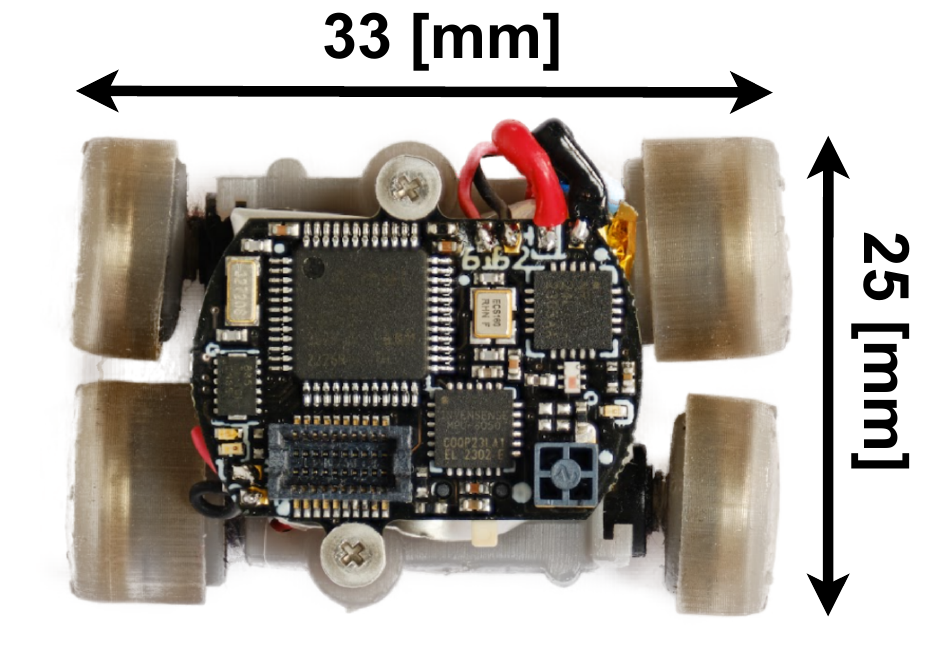}
    \footnotesize{(a) Real robot} \\ \vspace{1mm}
    \includegraphics[width=0.95\linewidth]{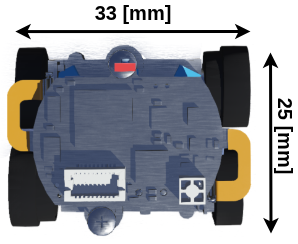}
    \footnotesize{(b) Simulated robot}
  \end{minipage}
  \hfill
  \begin{minipage}[b]{0.62\columnwidth} 
    \centering
    \includegraphics[width=\linewidth]{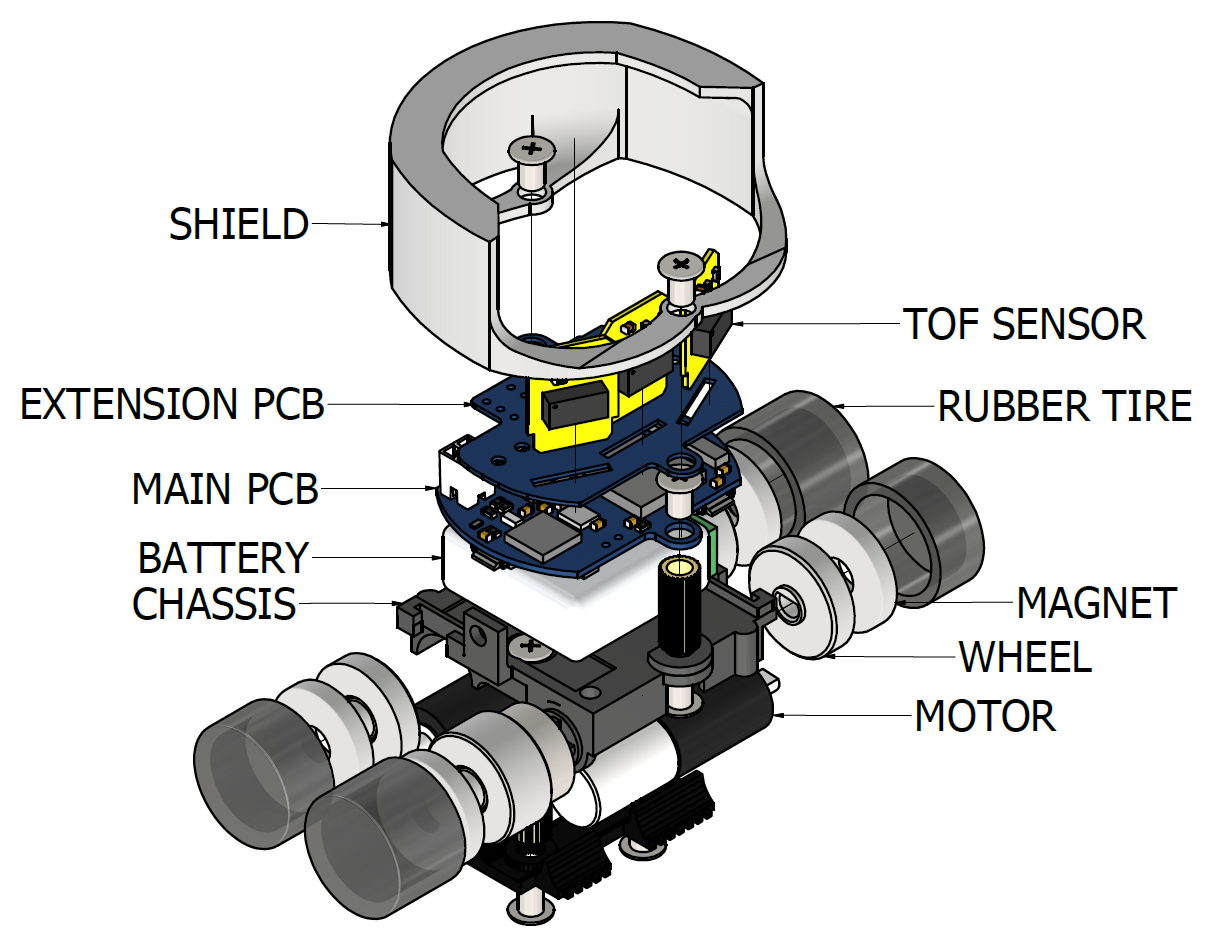}
    \footnotesize{(c) Robot CAD model}
  \end{minipage}
  \caption{The robot platform modeled in our simulation studies is shown. 
(a) Real and (b) simulated robot platform excluding its extension PCB. 
(c) CAD model of the robot indicating several key components: IMU (MPU6050), ToF sensors (VL53L0X), nRF24L01 radio, and SAMD21G18 MCU. 
We use calibrated sensor and actuator models to model realistic robot behavior.}
  \label{fig:three-in-one-column_robot}
\end{figure}
We develop a simulation framework leveraging the physics-based Webots robotic simulator \cite{WebotsWww.cyberbotics.com} and the Finite Element Analysis (FEA) software Abaqus \cite{DassaultSystemesWww.3ds.com/products/simulia/abaqus} to emulate the interaction between a swarm of miniaturized robots and a vibrating surface. A \(1\,\mathrm{m} \times 1\,\mathrm{m} \times 3\,\mathrm{mm}\) structure serves as the testbed. Vibration data is simulated in Abaqus on an evenly spaced grid and sampled in Webots at robot-specific locations using barycentric interpolation. A supervising node in Webots provides position data to the robots upon request.

\subsection{Miniaturized Robot Model}
\label{sec:robot-platform}
The simulated robot is modeled based on a real miniaturized platform shown in Figure~\ref{fig:three-in-one-column_robot}, derived from the design in \cite{Dementyev2016Rovables:Wearables}. It features an MPU6050 IMU for vibration sensing, an nRF24L01 radio module for communication, and a SAMD21G18 MCU for onboard processing. For collision avoidance, three VL53L0X Time-of-Flight (ToF) sensors are mounted on an external PCB. Robot locomotion on ferromagnetic surfaces is supported by magnetic wheels.

The simulation mirrors key hardware characteristics. The ToF beams are emulated from the robot’s base with the same field of view as the real sensors. The wheel encoder resolution is set to 12 counts per revolution. Accelerometer noise is modeled at \(60\,\mu g / \sqrt{\text{Hz}}\). Robot power constraints are also emulated, assuming a 100 mAh battery that supports roughly 30 minutes of typical sensing and actuation. Our on-going hardware development efforts are focused on upgrading the MCU to a SAMD51J20 and replacing the IMU with an industry-grade ISM330DHCXTR \cite{AmazonWww.aws.amazon.com/monitron/}, for our future studies.

\subsection{Surface Vibration Model}
\label{sec:data-generation}
We construct the surface model in Abaqus using S253JR construction steel with a Young’s modulus of \(E = 210\,\text{GPa}\) and Poisson’s ratio of \(\nu = 0.3\). We apply encastre boundary conditions to two opposing edges to represent fixed supports and inject ambient excitation by adding zero-mean Gaussian white noise accelerations, commonly used in SHM studies \cite{Mostafaei2025StatePerspectives}. This setup resembles a bridge section supported at both ends.
We discretize the structure into \(10\,\mathrm{mm} \times 10\,\mathrm{mm}\) linear shell elements, producing a \(101 \times 101\) grid. The dynamic response reveals two dominant mode shapes, \(\phi_1\) and \(\phi_2\), which account for 66\% and 29\% of the total vibrational energy, respectively. Areas where each mode shape exhibits zero vibrations correspond to zero-mode areas (see blue lines Figure \ref{fig:mode_shapes}).
We extract the mode shapes using frequency analysis and simulate time-domain vibrations by applying mode superposition in a dynamic analysis. Each scenario runs for 30 seconds and is sampled at 400 Hz, yielding approximately \(2.44 \times 10^8\) vibration samples per experimental scenario. To simulate damage, we reduce the thickness of shell elements in selected regions, lowering their structural stiffness and altering the vibrational response.

\section{Proposed Inspection Algorithm}
\label{sec:algorithm}
Each robot iteratively performs the following algorithm (Algorithm \ref{alg:damage_detection}) steps: data acquisition (Step 1), data processing (Step 2), uncertainty estimation (Step 3), and navigation (Step 4). Once the robot's estimation uncertainty drops below a predefined threshold, damage is estimated using the sum of mode shape curvatures (Step 5).
\subsubsection{Data Acquisition}
At the $k$th sampling location ${x_k}$, the robot stops and acquires acceleration data $a(t)$ for a duration of $T$ seconds.
Given the structure's natural frequencies of 17 Hz and 90 Hz, we choose a sampling rate of 400 Hz, which, according to the Nyquist criterion, allows accurate measurements up to 200 Hz.
The sampling duration $T$ is set to 15 seconds to provide a trade-off between frequency domain resolution, noise averaging, and robot battery life. 
\subsubsection{Data Processing} Following data-acquisition, $a(t)$ is processed to reduce data size and extract frequency content. Initially, a Hanning window is applied to the signal \( a(t) \) to minimize spectral leakage, resulting in the windowed signal $\hat a(t)$. The Fast Fourier Transform (FFT) is then applied to \( \hat{a}(t) \) to obtain its frequency content as $A(\omega) = \mathcal{FFT} \left\{ \hat{a}(t) \right\}.$
Using prior knowledge of undamaged natural frequencies \( f_i \), the frequency domain content \( \hat{A}_i(\omega) \) for each mode shape is extracted within a bandwidth $\Delta f$ around each \( f_i \) as:
\begin{subequations}
\label{eq:freq_portions}
\begin{align}
    \hat{A}_i(\omega) &= \left\{ {A}(\omega) \;\middle|\; 
    \frac{\omega}{2\pi} \in \left[f_i - \tfrac{\Delta f}{2},\, f_i + \tfrac{\Delta f}{2} \right] \right\}, \\
    \mathcal{A}_k &= \bigcup_{i=1}^{M} \hat{A}_i(\omega),
\end{align}
\end{subequations}
where $M$ equals the number of mode shapes considered ($M=2$ in our case). The union of frequency domain segments, denoted \(\mathcal{A}_k\), and the corresponding sampling location \(x_k\), is then broadcast to all other robots, which store it as: \(S \leftarrow S \cup \{(x_k, \mathcal{A}_k)\}\), thereby accumulating all locally acquired and received samples.
This approach reduces communication and data storage by focusing solely on frequency bands of size $|\hat{A}_i(\omega)|$ around each natural frequency, rather than considering the full spectrum size \(|A(\omega)|\), with data reduction proportional to \( (M \cdot |\hat{A}_i(\omega)|) /|A(\omega)| \).
\begin{figure}[t] 
    \centering
    \begin{minipage}{0.49\linewidth}
        \centering
        \includegraphics[width=\textwidth]{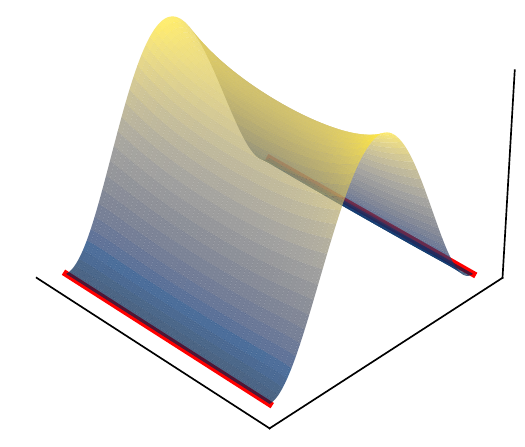} 
         \footnotesize{ (a) Vibration mode shape \(\phi_1\)}
        \label{fig:figure1}
    \end{minipage}
    \begin{minipage}{0.49\linewidth}
        \centering
        \includegraphics[width=\textwidth]{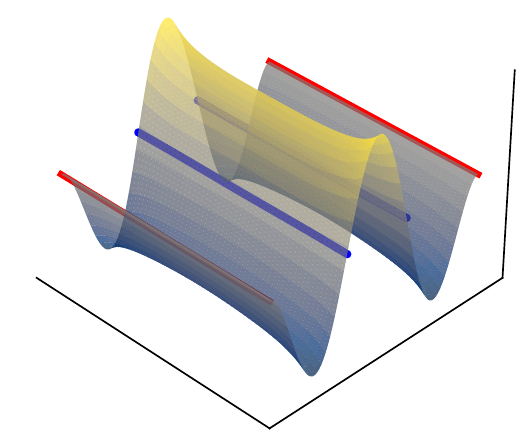} 
         \footnotesize{ (b) Vibration mode shape \(\phi_2\)}
        \label{fig:figure2}
    \end{minipage}
    \caption{Two dominant mode shapes of a \(1\,\mathrm{m} \times 1\,\mathrm{m} \times 0.003\,\mathrm{m}\) structure, clamped (encastrated) along two opposite edges, as highlighted in red. (a) The first mode shape \(\phi_1\) with natural frequency \(f_1 \approx 17\,\mathrm{Hz}\). (b) The second mode shape \(\phi_2\) with natural frequency \(f_2 \approx 90\,\mathrm{Hz}\). These mode shapes represent the vibration amplitude across the structure when excited at their respective natural frequencies. The zero-mode lines (blue lines) correspond to areas of no vibration, i.e. \(xy\)-plane intercepts.}
        \label{fig:mode_shapes}
\end{figure}
\subsubsection{Uncertainty Estimation}
The robot quantifies the estimation uncertainty around its sampling location $x_k$ using GP estimation \cite{Jang2020Multi-RobotProcess, Pillonetto2017DistributedApproximations, Mishra2018OnlineProcesses}. We adopt a simple decentralized GP method, laying groundwork for future developments. Assuming no prior information about the mode shape \(\phi_i\), we model $\phi_i$ as a zero-mean GP, that is,
\begin{equation}
    \label{eq:gp_process}
    \phi_i(x) \sim \mathcal{GP}\big(0, k(x, x')\big),
\end{equation}
where \(k(x, x')\) is the covariance function, for which we adopt the radial basis function (RBF) kernel as:
\begin{equation}
    \label{eq:rbf}
    k(x, x') = \sigma_v^2 \exp\left(-\frac{\| x - x' \|^2}{2l^2}\right),
\end{equation}
where the hyper parameters \(\sigma_v\) and \(l\) represent the signal's amplitude variation and length-scale (rate of decay), respectively. They can be learned from data by maximizing the marginal likelihood \cite{Rasmussen2004GaussianLearning}.
Given the sampling locations \(\mathbf{X} = \{x_1, \ldots, x_k\}\) and mode shape values \(\mathbf{\Phi} = \{\phi_i(x_1), \ldots, \phi_i(x_k)\}\), we compute the mode shape estimation $\hat \phi_i$ and corresponding uncertainty $\Sigma_i$ at an unobserved location \(x_*\) using the GP posterior distribution as:
\begin{equation}
    p(\phi_i(x_*) \mid \mathbf{X}, \mathbf{\Phi}, x_*) \sim \mathcal{N}(\hat{\phi}_i(x_*), \Sigma_i(x_*)),
\end{equation}
with predictive mean and uncertainty (variance) given as:
\begin{subequations}
\label{eq:predictionGP}
\begin{align}
    \hat{\phi}_i(x_*) &= K(x_*, \mathbf{X}) \left(K(\mathbf{X}, \mathbf{X}) + \sigma_n^2 I\right)^{-1} \mathbf{\Phi}, \\
    \Sigma_i(x_*) &= k(x_*, x_*) \nonumber \\
    &\quad - K(x_*, \mathbf{X}) \left(K(\mathbf{X}, \mathbf{X}) + \sigma_n^2 I\right)^{-1} K(\mathbf{X}, x_*),
\end{align}
\end{subequations}
where \(\sigma_n^2\) indicates the model uncertainty (induced by e.g. temperature, humidity, measurement noise etc.), and \(K(\cdot, \cdot)\) denotes the kernel matrix whose entries are computed using the RBF kernel in \eqref{eq:rbf}. Since \(\Sigma_i\) depends only on robot locations, uncertainty-based navigation can proceed without computing intermediate mode shape values $\mathbf{\Phi}$.
\subsubsection{Navigation}
The new target $x_{k+1}$ is selected using the uncertainty \(\Sigma_i\) around the robot's old location $x_k$. 
To this end, GP input samples $\mathbf{X}$ are taken within a radius $r_e$ of $x_k$ as $\mathbf{X}= \{x_j\in S| ||x_j - x_k|| < r_e\}$, and the target locations $\mathbf{X_*}$ are taken as evenly spaced points in polar coordinates around $x_k$ within a radius $r_t$ (where $r_t<r_e$) using $s_t$ radial and tangential steps.
Smaller \(r_e\) and \(s_t\) values reduce the size of \(\mathbf{X}\) and \(\mathbf{X_*}\), respectively, thereby lowering computational load and preventing the selection of targets which may be better suited for other robots.
\(x_{k+1}\) is now selected as the point in \(\mathbf{X_*}\) with maximum uncertainty above a threshold \(\theta_\Sigma\) and at least \(\theta_x\) away from all samples in \(S\), computed as:
\begin{align}
    \label{eq:navigation}
    x_{k+1} &= \arg\max_{x_* \in \mathbf{X_*}} \Sigma_i(x_*) \\
    \text{subject to} \quad &\Sigma_i(x_*) > \theta_\Sigma \quad \text{and} \quad \min_{(x_j, \mathcal{A}_j) \in S} \|x_* - x_j\| \geq \theta_x. \nonumber
\end{align}
If no target is found, the radii \( r_e \) and \( r_t \) are doubled and the process is repeated, similar to \cite{Jang2020Multi-RobotProcess}. If again no target is found, the robot proceeds to the Damage Estimation (Step (v)).
If $x_{k+1}$ exists, the robot moves towards the target by turning to the required heading and moving the required distance using the onboard gyroscope and wheel encoders.

To avoid multiple robots targeting the same \( x_{k+1} \), a randomized collision avoidance strategy is used: when the ToF sensors detect an obstacle within 1.5 robot body lengths, the robot turns by a random angle \( \sim \mathcal{U}(-\pi, \pi)\,\mathrm{rad} \) and moves a random distance \( \sim \mathcal{U}(0.05, 0.2) \,\mathrm{m}\), after which navigation resumes. If five collisions are triggered within a single navigation step, or if a received sample lies within a distance \( \theta_x \) of \( x_{k+1} \), the robot re-initiates navigation using its current position as the new starting point.

\begin{figure*}[t!]
    \centering
    \begin{minipage}[t]{0.59\textwidth} 
        \centering
        \includegraphics[width=\textwidth]{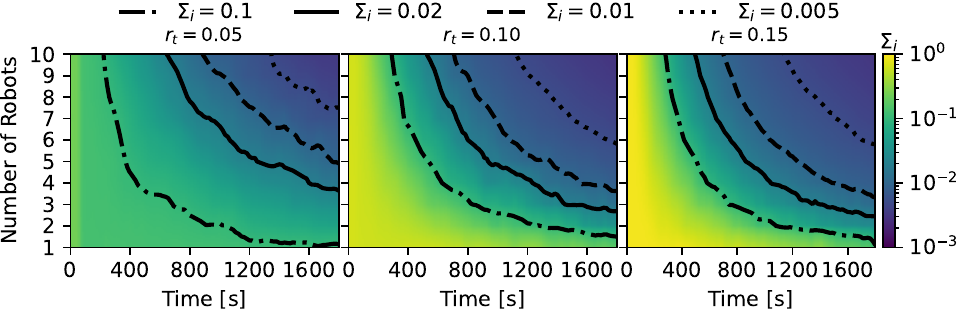}
                  \footnotesize{(a)}
    \end{minipage}%
    \hfill
    \begin{minipage}[t]{0.39\textwidth} 
        \centering
        \includegraphics[width=\textwidth]{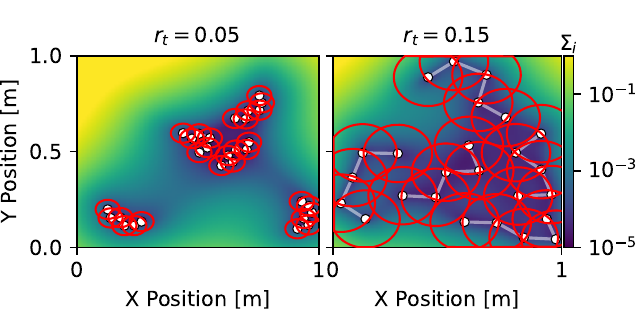}
                  \footnotesize{(b) }
    \end{minipage}%
\caption{
(a) Maximum observed GP variance $\Sigma_i$ over time for varying numbers of robots and exploration radii $r_t \in \{0.05, 0.10, 0.15\}\,\mathrm{m}$. Each heatmap shows the average GP variance observed by robots over time, with contour lines indicating thresholds ($\Sigma_i = 0.005$, $0.01$, $0.02$, $0.1$). Larger $r_t$ and swarm sizes more effectively reduce global uncertainty over time. In contrast, smaller $r_t$ values lead to faster local variance reduction (Time $<400$~s), but result in higher residual variance as time progresses. The optimal $r_t$ thus depends on both swarm and structure size.  
(b) Example variance maps at $T=100$ for two exploration radii using five robots. Left plot: $r_t = 0.05 \,\mathrm{m}$ results in local sampling and limited global coverage. Right plot: $r_t = 0.15\,\mathrm{m}$ promotes broader exploration and more uniform variance reduction. Robot trajectories are shown in white on both plots; red circles represent their exploration radii.}
    \label{fig:exploration_results}
\end{figure*}
\subsubsection{Damage Estimation}
Each robot derives a global damage estimate based on mode shape curvature analysis through combining FDD and GP estimation.
Assuming white noise excitations, the output spectral density around each natural frequency $w_i=2\pi f_i$ is approximated as:
 \begin{equation}
    \mathbf{G}_{yyi} (\omega) = \{ \hat{A}_i(\omega) \}\cdot \{ \hat{A}_i^*(\omega)\}^\top,
\end{equation}
where \(\hat{A}_i^*\) denotes the complex conjugate of \(\hat{A}_i\) \cite{Zimmerman2008AutomatedSystems, Brincker2001ModalDecomposition}. 
Provided $|S|=k$, the size of $\mathbf{G}_{yyi} $ equals $k\times k$.
The mode shape \(\phi_i\) is now approximated via SVD of \(\mathbf{G}_{yyi}(\omega)\) as:
\begin{equation}
    \mathbf{G}_{yyi}(\omega) = U_i Z_i U_i^\top,
\end{equation}
where \(U_i = [u_{i1}, \dots, u_{ik}]\) contains the unit-norm singular vectors (mode shapes), and \(Z_i = \mathrm{diag}(\lambda_{i1}, \dots, \lambda_{ik})\) contains the corresponding singular values (mode amplitudes) at the sampling locations. The (unitless) mode shape \(\phi_i\) is taken as the dominant singular vector \(u_{i1}\), with \(\|\phi_i\| = 1\).
Using \(\phi_i\), we derive the GP estimate \(\hat{\phi}_i\) across the structure with even spacing \(\Delta \phi\), setting the desired spatial resolution for damage detection. The input locations for this GP estimation equal \(\mathbf{X} = \{ x_j \mid (x_j, \mathcal{A}_k) \in S \}\), and the input values equal \(\mathbf{\Phi} = \{ \phi_i(x_j) \mid (x_j, \mathcal{A}_k) \in S \}\). Prior knowledge of the undamaged mode shape is used to tune the GP hyper parameters for each mode $\phi_i$. The curvature $\kappa_i$ is now approximated as \cite{Pandey1991DamageShapes}:
\begin{equation}
    \kappa_i(x_j) \approx \frac{\hat \phi_{i}(x_{j-1}) - 2 \hat \phi_{i}(x_{j}) + \hat \phi_{i}(x_{j+1})}{\Delta \phi ^2}.
\end{equation}
Comparing the curvature to undamaged data produces a damage estimate by summing over multiple curvatures as:
\begin{equation}
    D(x_j) = \sum_{i=1}^M w_i \left| \kappa_i(x_j) - \kappa_i^{\mathrm{prior}}(x_j) \right|,
    \quad \text{with} \quad \sum_{i=1}^M w_i = 1,
\end{equation}
where the weights \( w_i \) are set proportional to the contribution of each mode to the overall vibration response.
Finally, we define the damage score (z-score) at location \(x_j\) as:
\begin{equation}
    Z(x_j) = \frac{D(x_j) - \mu_D}{\sigma_D},
\end{equation}
where \(\mu_D\) and \(\sigma_D\) are the mean and standard deviation of $D$. A location is considered damaged if \(Z(x_j)\) exceeds the threshold \(\theta_Z\), which can be tuned for sensitivity.

\begin{algorithm}[t!]
\caption{Damage Detection Algorithm}
\label{alg:damage_detection}
\begin{algorithmic}[1]
\STATE Initialize dataset $S \leftarrow \emptyset$
\STATE Initialize target position $x_{k+1} \leftarrow x_0$ \COMMENT{ Initial position}
\STATE Initialize newLocationFound $\leftarrow $ True
\WHILE{newLocationFound}
    \STATE Navigate to $x_{k+1}$ (perform collision avoidance)
    \STATE $x_{k} \leftarrow x_{k+1}$
    \STATE Sample for $T$ seconds, i.e. $a(t)$ at $x_k$
    \STATE $\hat a(t) \leftarrow \text{Hanning}(a(t))$
    \STATE $A(\omega) \leftarrow \mathcal{FFT}(\hat a(t))$
    \STATE $\hat{A}_i(\omega) = \left\{ {A}(\omega) \;\middle|\; 
    \frac{\omega}{2\pi} \in \left[f_i - \tfrac{\Delta f}{2},\, f_i + \tfrac{\Delta f}{2} \right] \right\}$
    \STATE $  \mathcal{A}_k \leftarrow \bigcup_{i=1}^{M} \hat{A}_i(\omega)$
    \STATE $S \leftarrow S \cup \{(x_k,  \mathcal{A}_k)\}$
    \STATE newLocationFound $\leftarrow$ False
    \STATE $i\leftarrow 1$
    \WHILE{!newLocationFound \text{ AND } $i\leq 2$}
    \STATE $\mathbf{X_*} = \left\{ (r, \theta) \middle| r = \tfrac{imr_t}{s_t}, \theta = \tfrac{2\pi n}{s_t}, m,n = 1\dots s_t \right\}$
    \STATE $\mathbf{X} = \{x_j \in S\  |\  ||x_j - x_k|| < i\cdot r_e\}$
    \STATE Select target $x_{k+1}$ from $\mathbf{X_*}$ (Equation \eqref{eq:navigation})
     \IF{exists $x_{k+1}$}
        \STATE newLocationFound $\leftarrow$ True
    \ELSE
        \STATE $i \leftarrow i +1$
    \ENDIF
    \ENDWHILE
\ENDWHILE
\STATE Apply FDD on \(S\) to estimate \(\phi_i(x_j)\), for \(i = 1, \ldots, M\)
\STATE Extract sampling locations \(\mathbf{X} = \{ x_j \mid (x_j,  \mathcal{A}_k) \in S \}\)
\STATE Extract mode values \(\mathbf{\Phi} = \{ \phi_i(x_j) \mid (x_j,  \mathcal{A}_k) \in S \}\)
\STATE Compute evenly spaced $\hat{\phi_i}$ using GP for $i=1,...,M$
\STATE Compute damage estimate $D(x_j)$ (sum of curvatures)
\STATE Indicate damaged regions using z-score
\end{algorithmic}
\end{algorithm}


\section{Experiments and Results}
\label{sec:experiments-results}
We assess our proposed inspection approach through a three-stage experiment on the \(1\,\mathrm{m} \times 1\,\mathrm{m} \times 0.003\,\mathrm{m}\) surface structure illustrated in Figure~\ref{fig:mode_shapes}.  
First, we evaluate the impact of different radii \(r_t\) and swarm sizes on estimation uncertainty, and determine suitable values for these two parameters.  
Second, we optimize the GP hyper parameters ($\sigma_v,\sigma_n$ and $l$) for the two mode shapes of the given surface structure to improve damage estimation accuracy.  
Third, we assess our approach's performance across 10 damage scenarios, where a swarm of five robots is tasked with detecting and localizing randomized and unknown damages.

\subsubsection{Stage I - Exploration}
We study the effect of exploration by varying \(r_t \in \{0.05, 0.10, 0.15\}\,\mathrm{m}\) in swarms of 1 to 10 robots.  
Each robot selects target sampling locations within the area defined by \(r_t\), evenly discretized using \(s_t = 10\) steps.  
The encircling estimation radius is set to \(r_e = r_t + 0.1\,\mathrm{m}\) to avoid boundary inaccuracies.
When applicable, available hardware may impose upper limits on $r_e$ and $s_t$.
We consider a minimal sample distance of $\theta_x = 0.01 \,\mathrm{m}$ to avoid duplicate samples.
Within the estimation, we fix the GP hyper parameters to \(l = 0.1\,\mathrm{m}\), \(\sigma_v = 1\), and \(\sigma_n = 0.01\). Since these parameters mainly affect the scale of \(\Sigma_i\) rather than the peak locations, their exact values are less critical, and fixing them simplifies exploration.  
The resulting uncertainty profiles are displayed in Figure~\ref{fig:exploration_results}a and show lower observed uncertainty at early times (\(\text{Time} < 400\,\mathrm{s}\)) for smaller radii (\(r_t = 0.05\,\mathrm{m}\)), while larger radii (\(r_t = 0.15\,\mathrm{m}\)) become more effective at reducing uncertainty as time progresses. In particular, small swarm sizes benefit from larger \(r_t\) values to effectively reduce global uncertainty within limited battery life.  
Figure~\ref{fig:exploration_results}b illustrates a 5-robot example, showing that smaller \(r_t\) values reduce uncertainty locally, while larger \(r_t\) values enhance global coverage.
Thus, the value of \(r_t\) should be set proportionally to the surface structure size and the swarm size.
Correspondingly, we select a swarm size of 5 robots using $r_t = 0.15\,\mathrm{m}$ for following experiment stages.

\subsubsection{Stage II - Optimization}
We optimize the 
unitless variables \(\sigma_v\), \(\sigma_n\), and length scale \(l\) for each mode shape \(\phi_1\) and \(\phi_2\) using an implementation provided by \texttt{scikit-learn}~\cite{Scikit-learn2024GaussianProcesses}, which returns optimal parameters by maximizing the marginal likelihood~\cite{Rasmussen2004GaussianLearning}.
We run 10 simulations with undamaged vibration signals and acquire samples using a threshold \(\theta_{\Sigma} = 0.025\). Mode shape values at the sample locations, estimated via FDD, serve as input to the optimization, which is run for 20 iterations. 
The results in Table~\ref{tab:gp-params} indicate higher signal amplitudes \((\sigma_v)\) and longer length scales \((l)\) for \(\phi_1\). This difference is expected since \(\phi_1\) is less curved (see Figure~\ref{fig:mode_shapes}), 
causing its amplitudes to change more gradually.
We halve each \( l \) to enhance sensitivity to locally induced damage variations.
An excessive reduction may require adjustment of \(\sigma_v\) and \(\sigma_n\) to mitigate corresponding noise amplifications.

\begin{table}[b!]
\centering
\caption{Optimized GP parameters for \(\phi_1\) and \(\phi_2\)
}
\label{tab:gp-params}
\begin{tabular}{lccc}
\hline
\textbf{Parameter} &\textbf{Mode $\phi_1$}& \textbf{Mode $\phi_2$} & \textbf{Bounds [min, max]} \\
\hline
\(\sigma_v\) [unitless]  & 104.4 &40.51 & [\( 10^{-10}\), \( 10^{10}\)] \\
\(\sigma_n\) [unitless] & \(4.82 \cdot 10^{-5}\) &\(4.56 \cdot 10^{-5}\)& [\( 10^{-10}\), \(10^{10}\)] \\
\(l\) [m]         & 0.245 & 0.149& [\( 10^{-10}\), \( 10^{10}\)] \\
\hline
\end{tabular}
\end{table}

\subsubsection{Stage III - Damage Detection}
We validate our damage detection method through 100 randomized simulations for each of 10 damage scenarios. In each scenario, the number of damages is sampled from \(\mathcal{U}(1, 3)\), the surface area from \(\mathcal{U}(2500, 25000)\,\mathrm{mm}^2\), and a single uniform thickness removal from \(\mathcal{U}(0.5, 1)\,\mathrm{mm}\). Damage shapes are formed by selecting a random start element and stochastically adding neighboring elements until the damage size is met.  
Randomization is achieved through initial positions, seeds, and by selecting a randomized vibration segment of 15 second from the 30 seconds available in Abaqus.
We reduce communication load by selecting \(\Delta f = 10\,\mathrm{Hz}\), achieving a data reduction factor of \((M \cdot |\hat{A}_i(\omega)|)/|A(\omega)| = (2 \cdot 300)/3000 = 0.2\) relative to the full bandwidth. The remaining parameters follow the same configuration as outlined in Stages I and II.
\begin{figure}[t!]
    \centering
    \includegraphics[width=0.85\columnwidth]{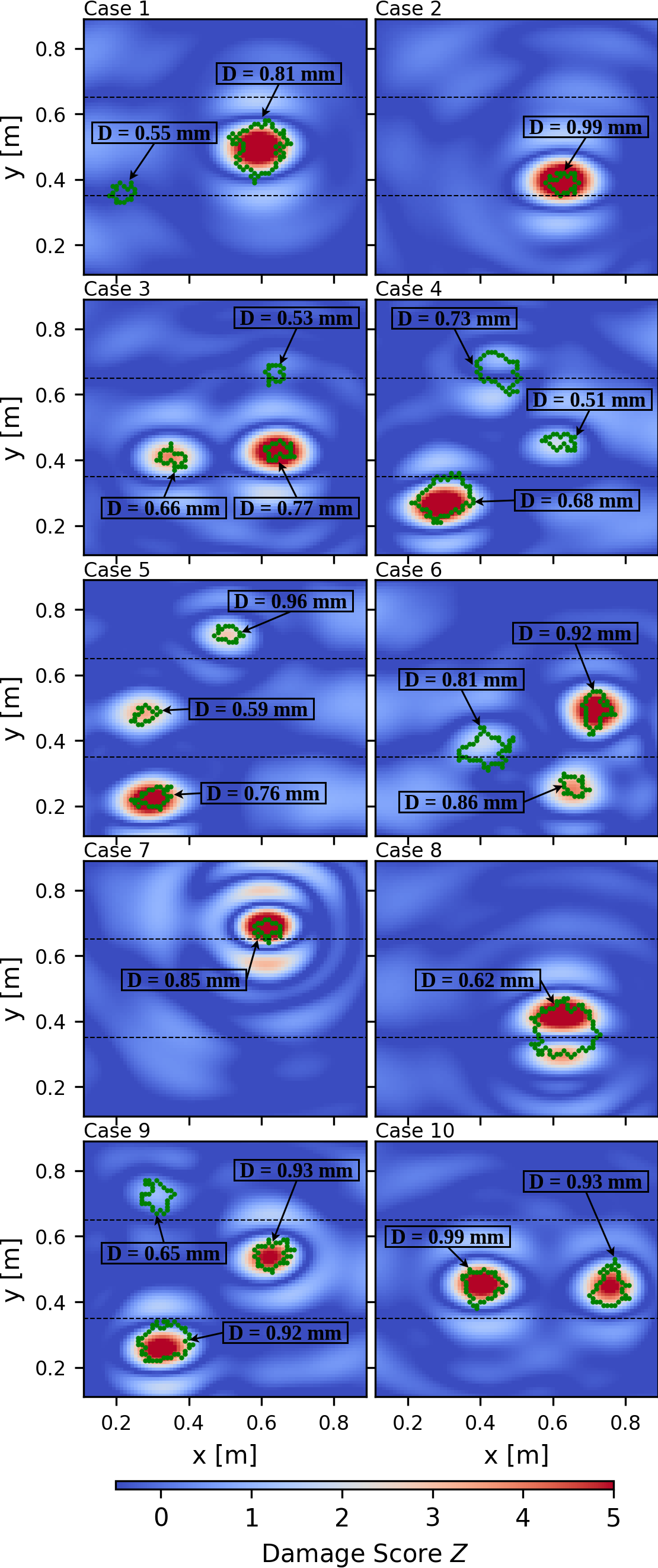}
    \caption{Average damage scores computed over robots and 100 experiments per damage case. The green regions outline the true damage present on the surface, with "D" corresponding to the thickness of material removed. The plots show that larger damage areas are reconstructed more accurately than smaller ones. Additionally, damage near the zero-mode lines (dotted black) is difficult to detect due to minimal vibration in these regions (e.g. Case 1 and 4). Overall, most of the surface is classified correctly, and false positive values remain in the vicinity of true damage.}
    \label{fig:damage_results}
\end{figure}

The results in Figure~\ref{fig:damage_results} demonstrate effective damage detection in most cases (e.g., Cases 2, 5, and 10), while Cases 4, 6, and 9 exhibit damages with ambiguous damage scores in damaged areas. We attribute this to two factors: in Case 9, the other, more severe, damages may influence the damage score's (z-score) mean and standard deviation resulting in less severe damages getting overshadowed; and (ii) in Case 4 and 6, damages located near zero-mode areas shown by dotted black lines have reduced observability due to minimal surface vibration amplitudes (see Figure~\ref{fig:mode_shapes}). 
This is also observed in Case 1 and 3, where the undetected damages are located within the vicinity of zero mode areas. 
Due to the minimal structural response in these areas, damages are harder to detect and more susceptible to measurement noise.
One possible solution is to consider higher order mode shapes, aiming to increase the vibration amplitudes in zero mode regions. 
We further quantify our approach's performance through accuracy, precision, recall, and number of detected true damages by thresholding damage scores at $\theta_Z = 2$ and comparing them to ground truth elements at $\Delta \phi = 10 \,\mathrm{mm}$ spacing. The results in Table~\ref{tab:accuracy_metrics} show the overall accuracy is high (97\%), indicating that most of the surface elements are correctly classified. 
The recall indicates that $73\%$ of the actual damage is detected, and cases exhibiting lower recall mainly involve damage located near zero-mode lines (e.g., Case 1, 3, 4, and 8). This is also reflected in the number of detected true damages.
The precision scores reveal that not all detected damage corresponds to true damage. However, from Figure~\ref{fig:damage_results} we conclude that false positives tend to cluster near actual damage, while distant misclassifications are rare.
Our findings suggest that the proposed method can effectively localize structural damage, though capturing the exact size and shape remains challenging. Moreover, performance depends not only on robot behavior but also on the structure’s characteristics like the zero mode areas, which must be considered carefully in performance assessments.

\section{Conclusion and Future Work}
\label{sec:discussion-conclusion}
We present a distributed structural damage detection approach using a swarm of miniaturized, mobile, vibration-sensing robots.
Our method combines GP-based exploration with established SHM techniques for estimating structural mode shape curvature.
To reduce communication overhead and computational cost, the approach leverages limited-bandwidth FDD and reduced-radius exploration, enabling practical deployment on resource-constrained robots.
Our findings indicate that an effective balance between swarm size and exploration radius is key to achieving comprehensive coverage. The system achieves high accuracy, localizing 73\% of damage instances in the evaluated scenarios.
Limitations include reduced sensitivity in regions with low structural response and false positives occurring near true damage sites.
Our contributions are twofold:
(i) the integration of SHM techniques with robotic, GP-driven exploration and estimation, bridging two previously disconnected domains; and
(ii) a realistic simulation framework to support future research on vibration-based inspection using robot swarms.

Future work will focus on reducing communication and computational demands by employing sub-swarms for localized search and leveraging local instead of global structural mode shape estimations. Additionally, the current method relies on FDD for mode shape estimation, while GPs can offer multidimensional modeling, enabling damage detection directly using vibration signals. Ongoing efforts aim to integrate these enhancements and validate the approach via real-world experiments using the robot presented here.

\begin{table}[h!]
\centering
\caption{Detected damages and scores (mean$\pm$std) per damage case 
}
\label{tab:accuracy_metrics}
\begin{tabular}{lcccc}
\hline
\textbf{Case} & \textbf{Acc. (\%)} & \textbf{Prec. (\%)} & \textbf{Rec. (\%)} & \textbf{Detected} \\
\hline
Case 1 & $98 \pm 0$ & $78 \pm 3$ & $75 \pm 1$ & $1/2$ \\
Case 2 & $97 \pm 0$ & $22 \pm 3$ & $100 \pm 0$ & $1/1$ \\
Case 3 & $96 \pm 1$ & $26 \pm 4$ & $66 \pm 11$ & $2/3$ \\
Case 4 & $96 \pm 1$ & $61 \pm 8$ & $42 \pm 4$ & $1/3$ \\
Case 5 & $97 \pm 1$ & $34 \pm 4$ & $81 \pm 12$ & $2/3$ \\
Case 6 & $96 \pm 0$ & $42 \pm 4$ & $51 \pm 6$ & $2/3$ \\
Case 7 & $96 \pm 1$ & $14 \pm 3$ & $97 \pm 5$ & $1/1$ \\
Case 8 & $97 \pm 1$ & $56 \pm 6$ & $65 \pm 3$ & $1/1$ \\
Case 9 & $96 \pm 1$ & $59 \pm 8$ & $59 \pm 5$ & $2/3$ \\
Case 10 & $98 \pm 0$ & $61 \pm 5$ & $92 \pm 4$ & $2/2$ \\
\hline
\textbf{Overall} & \textbf{$97 \pm 1$} & \textbf{$45 \pm 20$} & \textbf{$73 \pm 19$} & \textbf{$15/22$} \\
\hline
\end{tabular}
\end{table}
\bibliographystyle{ieeetr}
\bibliography{references}



\end{document}